\documentclass[10pt,twocolumn,letterpaper]{article}

\usepackage{cvpr}              

\usepackage{amsmath,amsfonts,bm}

\def\eqref#1{equation~\ref{#1}}

\def\1{\bm{1}}

\DeclareMathAlphabet{\mathsfit}{\encodingdefault}{\sfdefault}{m}{sl}
\SetMathAlphabet{\mathsfit}{bold}{\encodingdefault}{\sfdefault}{bx}{n}

\usepackage{wrapfig}
\usepackage{hyperref}
\usepackage{url}
\usepackage{graphicx}
\usepackage{amsmath}
\usepackage{amssymb}
\usepackage{nth}
\usepackage{booktabs}
\usepackage{times}
\usepackage{epsfig}
\usepackage{graphicx}
\usepackage{amsmath}
\usepackage{amssymb}
\usepackage{stfloats}
\usepackage{multirow}	
\usepackage{float}
\usepackage{xcolor}
\usepackage{hyperref}
\usepackage[accsupp]{axessibility}
\hypersetup{
    colorlinks=true,
    linkcolor=blue,
    filecolor=magenta,      
    urlcolor=cyan,
    pdftitle={Overleaf Example},
    pdfpagemode=FullScreen,
    }

\definecolor{ForestGreen}{RGB}{34,139,34}

\title{Therbligs in Action: Video Understanding through Motion Primitives}

\def\cvprPaperID{*****} 
\def\confName{CVPR}
\def\confYear{2023}

\begin{document}

\title{Therbligs in Action: Video Understanding through Motion Primitives}

\author{Eadom Dessalene, Michael Maynord, Cornelia Ferm\"uller, Yiannis Aloimonos\\
University of Maryland, College Park\\
College Park, MD 20742, USA\\
{\tt\small \{edessale,maynord,fermulcm,jyaloimo@umd.edu\}}
}
\maketitle

\begin{abstract}
In this paper we introduce a rule-based, compositional, and hierarchical modeling of action using Therbligs as our atoms. Introducing these atoms provides us with a  consistent, expressive, contact-centered representation of action. Over the atoms we introduce a differentiable method of rule-based reasoning to regularize for logical consistency. Our approach is complementary to other approaches in that the Therblig-based representations produced by our architecture augment rather than replace existing architectures' representations. We release the first Therblig-centered annotations over two popular video datasets - EPIC Kitchens 100 and 50-Salads. We also broadly demonstrate benefits to adopting Therblig representations through evaluation on the following tasks: action segmentation, action anticipation, and action recognition - observing an average 10.5\%/7.53\%/6.5\% relative improvement, respectively, over EPIC Kitchens and an average 8.9\%/6.63\%/4.8\% relative improvement, respectively, over 50 Salads. Code and data will be made publicly available.
\end{abstract}
\section{Introduction}

\begin{figure}[b!]
\begin{center}
  \includegraphics[width=.4\textwidth]{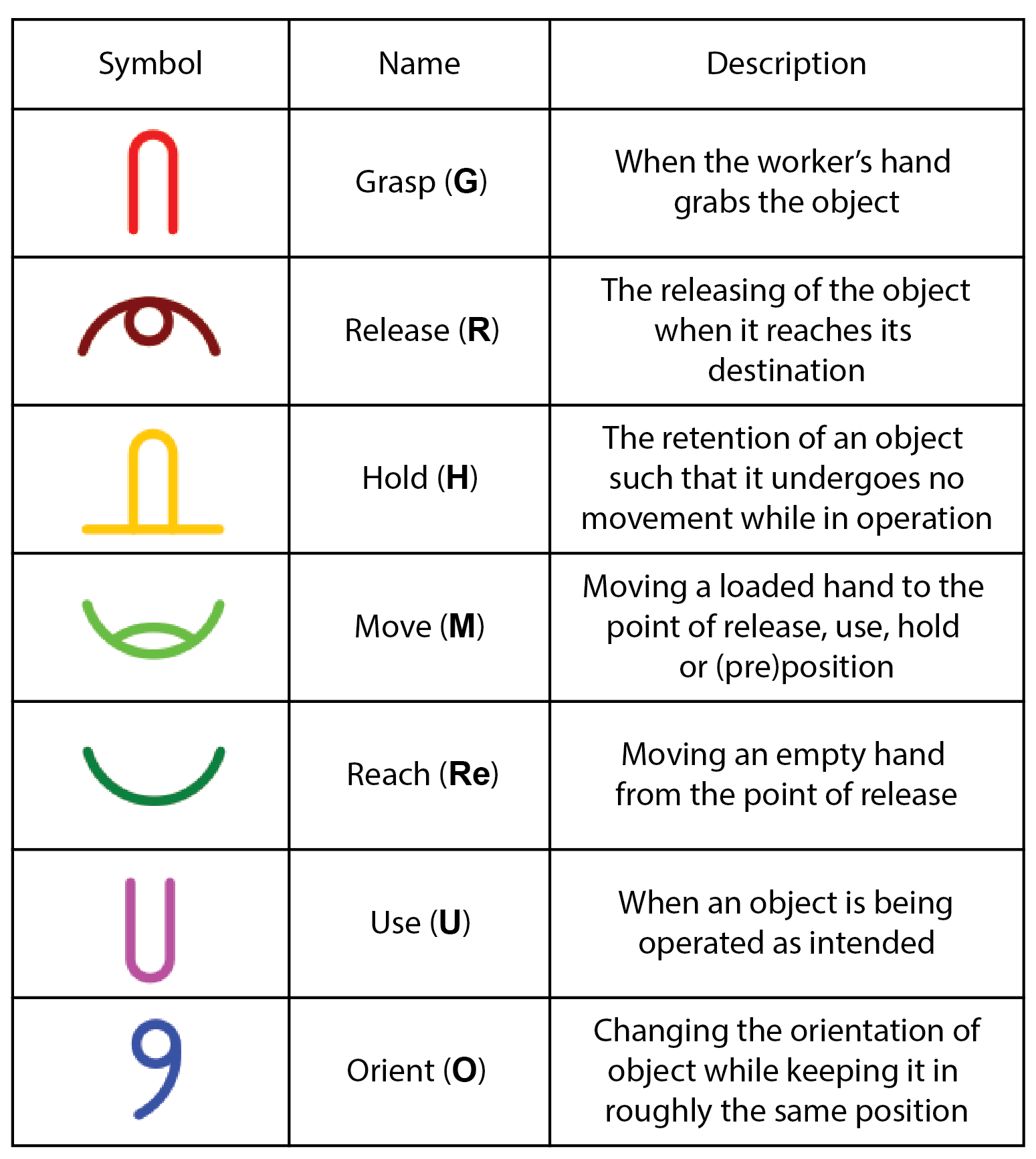}
 \end{center}
  \caption{Listed above are the Therbligs we select, their symbolic illustrations as introduced by Gilbreths, and brief descriptions of their usage.}
  \label{figure:therbligs}
\end{figure}









\begin{figure*}[t!]
\begin{center}
  \includegraphics[width=.7\textwidth]{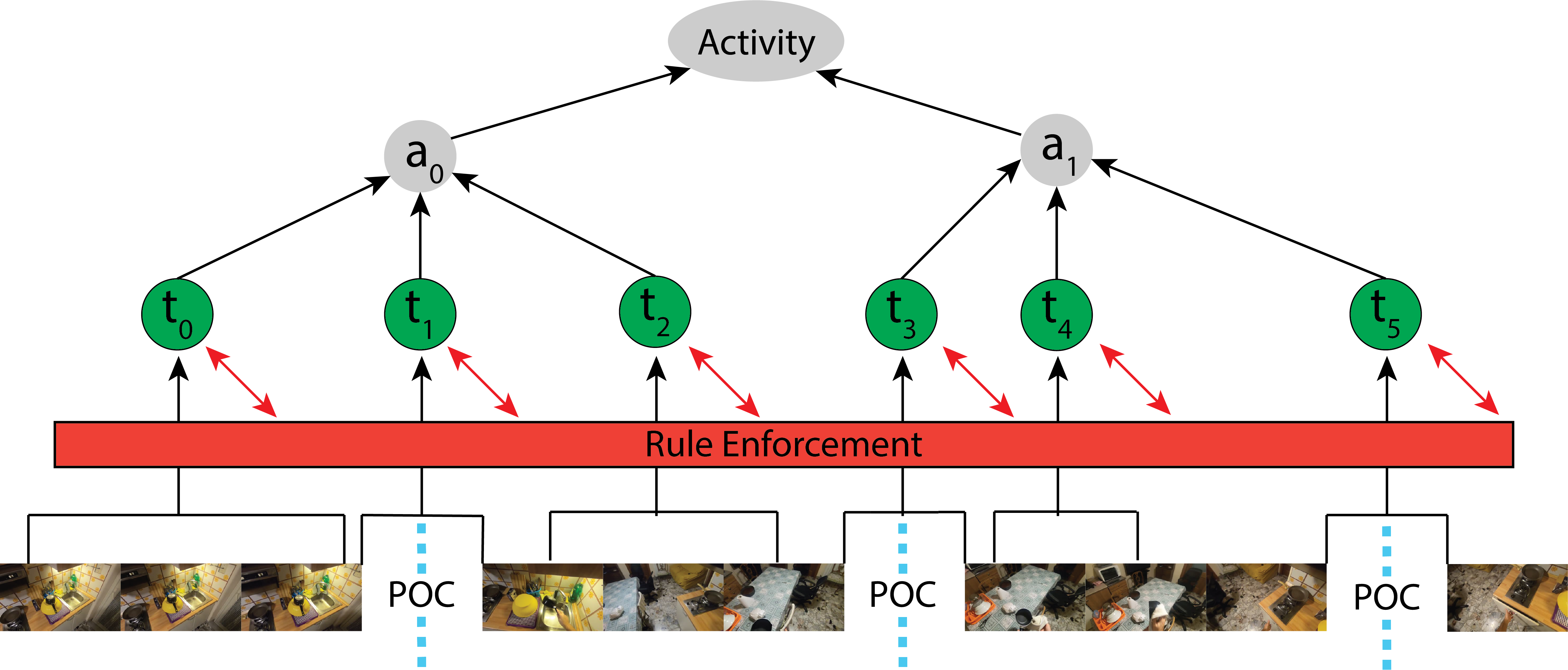}
  \end{center}
\vspace{-1.5em}
    \caption{We introduce the use of {\color{ForestGreen} Therbligs ($t_i$)} in video understanding as a consistent, expressive, symbolic representation of sub-action. Points of Contact (indicated by the {\color{blue} divider dashes}) are necessarily associated with Therbligs and/or their boundaries. Because of the unambiguity of Points of Contact, Therblig boundaries gain precision and are non-overlapping. On top of Therblig atoms we construct a framework for {\color{red} Rule Enforcement}, enforcing greater logical consistency through commonsense rules. This rule-based framework allows for the easy introduction of long-term constraints. Therblig atoms are then composable into actions ($a_i$), which are in turn composable into activities.
    }
    \label{fig:hierarchy}
    
\end{figure*}


We propose the use of \textit{Therbligs} - a low-level mutually exclusive contact demarcated set of sub-actions. These Therbligs are \textbf{consistent} in that a given action segment has only a single Therblig representation, and Therbligs are \textbf{expressive} in that they capture the meaningful physical aspects of action relevant to action modeling. Therbligs were introduced in the early \nth{20} century as a set of $18$ elemental motions used to analyze complex movement - see the Supplementary Materials for a brief historical background. We adopt $7$ Therbligs pertaining to those involving the manipulation of objects. See Figure \ref{figure:therbligs} for our Therblig set.



The benefits of our Therblig-centered framework include: compositionality \& hierarchy; rule-based reasoning; resolution of semantic ambiguity; contact-centered precision of temporal boundaries of action. 

Contact transitions demarcate Therblig boundaries, giving Therbligs a consistency which methods relying on annotators' intuited demarcations lack \footnote{See \cite{alwassel2018diagnosing} for a case study on how annotators have difficulties coming to a consensus on when actions begin and end.}. Between points of contact exist contact states represented by a binary class (contact, no contact) for each object present, which are wholly captured by Therbligs. As objects in contact are the primary objects of interaction and define the space of possible actions, they provide meaningful information for the modeling of action. 

Therblig atoms are then composable into higher entities, including full actions. These actions are in turn composable into sequences constituting activities. We then have the hierarchy of representation illustrated in Figure \ref{fig:hierarchy}. At the lowest, and instantaneous, level are points of contact, between which exist Therbligs with temporal extension, on top of which exist action, permutations of which constitute longer activities.


Architectures built upon Therbligs for the modeling of action gain temporal precision through points of contact as well as meaningful information captured by contact states. Therbligs also exhibit semantic mutual exclusivity in that there is one and only one Therblig interpretation of a sequence, as opposed to the many interpretations when action labels are intuited \footnote{Some additional structure is needed for complete mutual exclusivity - see the Supplementary Materials for discussion on this structure.}, leading to semantic ambiguity. As a consequence of Therbligs, semantic ambiguity at the action-level is constrained by the deeper grounding of action labels in explicit action dynamics (see Figure \ref{fig:hierarchy}).

Unlike higher level actions, Therbligs enable the imposing of a contact-based logic defining preconditions and postconditions in the form of states of contact before and after Therbligs. For example, an object being \textit{moved} must be preceded by \textit{grasp} and proceeded by a \textit{release}. The rules of this logic interface at the Therblig level of the hierarchy. These rules allow for biasing towards consistency between contact states and Therblig predictions within a loss term (see Section \ref{section:diff}), and provide constraints over possible action sequences (see Section \ref{section:explicit}).

\begin{figure*}[t!]
  \centering
  \includegraphics[width=.75\textwidth]{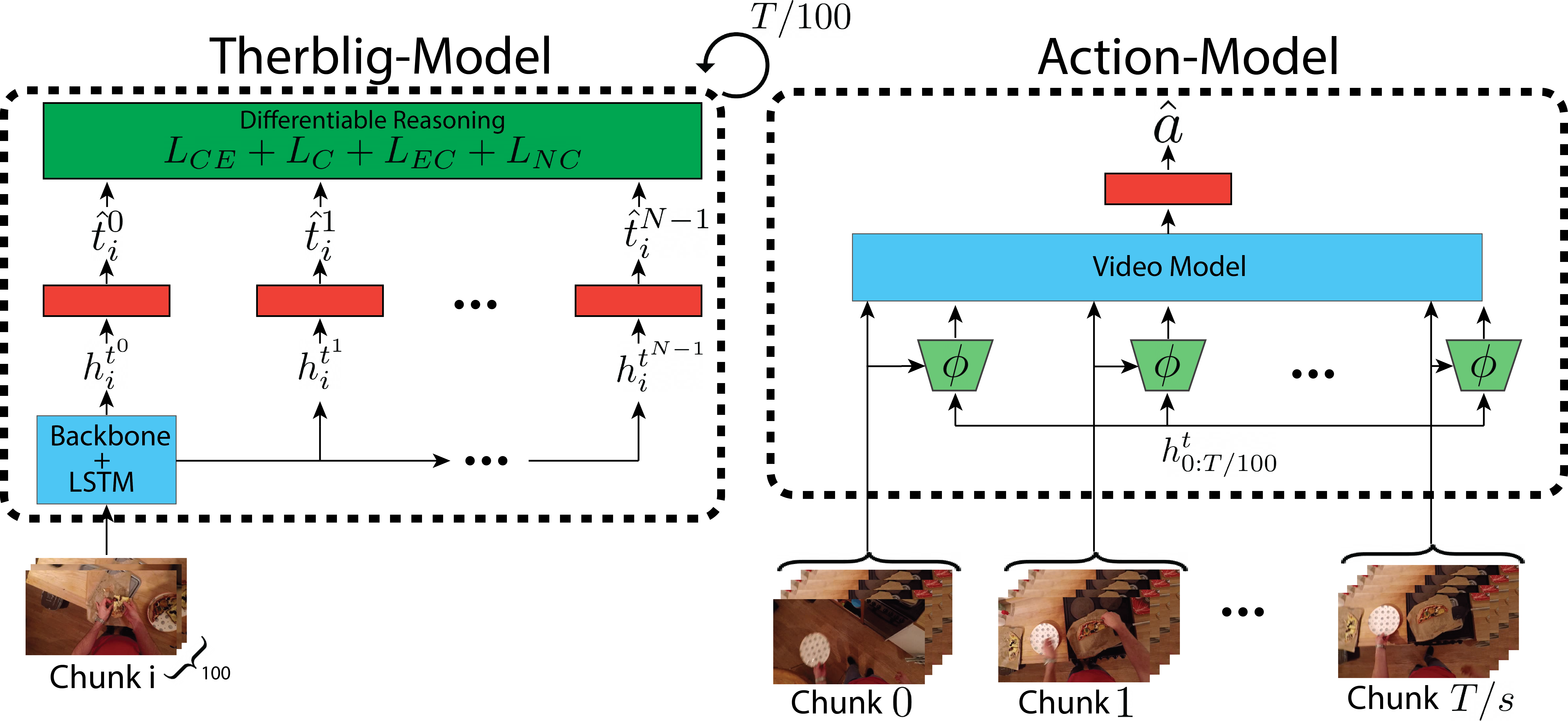}
 \caption{Architectural diagram of our framework. \textbf{Therblig-Model} takes a stack of $K=100$ frames as input, feeding them to various backbone video architectures followed by a 2-layer GRU ({\color{blue} Backbone + LSTM}), which in turn produces hidden states $h_i^t$, passing through all $T/100$ stacks of the video. Hidden states $h_i^t$ are fed to {\color{red} fully connected} layers, followed by a Gumbel-Softmax operation, producing Therblig predictions amenable to differentiable reasoning. \textbf{Action-Model} takes a sliding window with window size $W$ over the original video sequence with stride $s$, both values depending on the choice of architecture. These windows are fed to $\phi$, an attention mechanism consisting of a $2$ layer MLP - this MLP attends over the hidden states produced by \textbf{Therblig-Model}. The blended features produced by $\phi$ are fed together with the video window to a ({\color{blue} Video Network}) predicting action class likelihood $a$. See Sections \ref{section:Therblig-Model} and \ref{section:actionlstm} for details.}
 \label{fig:architecture}
\end{figure*}

In producing sub-action level symbolic representations, our proposed hierarchical architecture is comprised of two main components; the \textbf{Therblig-Model}, which maps video to Therbligs; and, the \textbf{Action-Model}, which maps video and Therbligs to actions. The Therblig-Model is optimized over a loss including structure-aware terms for contact consistency and Therblig consistency by incorporating differentiable reasoning. Figure \ref{fig:architecture} illustrates our architecture. This architecture is complementary to, rather than in competition with, existing architectures for action modeling - Therblig representations can be easily integrated through concatenation with existing feature representations. We demonstrate this with two state-of-the-art approaches to action segmentation - MSTCN++ \cite{li2020ms} and ASFormer \cite{yi2021asformer} and four popular approaches to action recognition - I3D \cite{carreira2017quo}, ViViT \cite{arnab2021vivit}, TimeSFormer\cite{bertasius2021space}, and MoViNet\cite{kondratyuk2021movinets}.







We evaluate our approach over the tasks of action segmentation, action recognition, and action anticipation. We evaluate over the EPIC Kitchens 100 and 50-Salads datasets.


The primary contributions of our work are as follows:
\begin{itemize}
    \item Therbligs, a consistent, expressive symbolic representation of sub-action centered on contact. 
    \item Rules: Flexible and differentiable constraining of intuitive constraints on arrangement of atomic actions, informed by commonsense rules.
    \item Novel hierarchical architecture composed of a Therblig-Model and Action-Model. Representations produced by the Therblig-Model can be easily integrated into other approaches, as we demonstrate with six popular action understanding approaches. 
    \item Dataset: We release the first Therblig-centered annotations over two popular video datasets. 
\end{itemize}

The rest of this paper is structured as follows: Section \ref{related} discusses related works, Section \ref{methods} introduces our proposed method, Section \ref{experiments} describes the experiments, in Section \ref{Discussion} we provide discussion and in Section \ref{conclusion} we conclude.

\section{Related Works}
\label{related}
\subsection{Sub-Action Video Datasets}
There exist several datasets that provide sub-action level annotations as a means of resolving semantic and temporal ambiguity in annotation, and enabling the hierarchical modeling of action \cite{ji2020action,shao2020finegym,shao2020intra}. FineGym \cite{shao2020finegym} introduces fine-grained action annotations for actions in gymnastics, but suffers from a difficult and expensive data collection process. TAPOS \cite{shao2020intra} manually breaks actions into sub-actions for Olympics videos via temporal action parsing. Other datasets producing sub-action level annotations are of an instructional nature \cite{kuehne2014language,rohrbach15ijcv,stein2013combining}, providing annotations for steps of various cooking activities. Our Therblig annotations differ from other sub-action ontologies by 1) resolving temporal ambiguity by means of contact, 2) having a simple, logically consistent data collection process enabled through the imposing of commonsense rules, and 3) being flexible in application to a wide variety of datasets within the realm of object manipulation without relying on domain expertise.

\subsection{Contact in Video Understanding}
Contact has proven to be a useful feature in several tasks of interest in computer vision, such as hand-object pose estimation \cite{cao2021reconstructing,karunratanakul2020grasping,shan2020understanding}, character animation, kinematic pose estimation \cite{rempe2020contact}, etc. However, the vast majority of contact-centered approaches apply in the single-image setting, and few works \cite{ji2020action} consider the modeling of contact for use in action understanding, despite contact being a defining characteristic of all physical human interaction. Ego-OMG \cite{dessalene2020egocentric} approaches the task of action anticipation, representing long sequences of manipulation activity through sequences of discrete states, each state delineated by the making and breaking of contact. Rather than directly extract contact from each video frame as in \cite{dessalene2021forecasting,dessalene2020egocentric}, we instead propose the adoption of Therbligs, a contact-centered representation governed by simple contact-based rules.

\section{Methods}
\label{methods}




For video segment $s_i$ with $T$ frames, the Therblig representation $t_i$ is a sequence of $N$ Therblig tuples for every $100$-frame chunk of $s_i$, and the contact representation $c_i$ represents the objects in contact at the end of that video segment. Each Therblig annotation $t_i$ is a sequence of the form $(v_0, o_0), ... (v_{N-1}, o_{N-1})$, where $v_j \in V$ and $V = \{\emptyset, Re, M, G, R, U, O, H\}$. In other words, $v_j$ indicates the Therblig verb and each $o_j$ indicates the object of interaction. See Figure \ref{figure:therbligs} for full descriptions of each element of $V$ (not included is $\emptyset$, corresponding to an empty sequence where no Therbligs occur). To illustrate, see the \textit{Therbligs} and \textit{Contacts} rows for annotations $t_i$ and $c_i$, respectively, in Figure \ref{fig:qualitative}. We set the maximum number of possible Therblig annotations per sequence $N$ to $6$. Each contact annotation $c_i$ is a tuple of the form $(c_i^r, c_i^l)$, where $c_i^r$ corresponds to the class of the object held by the right hand, and $c_i^l$ corresponds to the class of the object held by the left hand.

Given video segment $s_i$, our goal is to infer the relevant action class likelihood over the tasks of action recognition and anticipation, and to infer the relevant action class likelihood for each frame over the task of action segmentation. We do this by means of a novel hierarchical architecture as described in Subsection \ref{section:architecture}. This architecture consists of two levels; a Therblig-Model (\ref{section:Therblig-Model}) and an Action-Model (\ref{section:actionlstm}). We then describe our rule-based reasoning formulation in Subsection \ref{section:training}, and detail the Therblig annotation collection process in Subsection \ref{section:crowdsourcing}.



\subsection{Architecture}
\label{section:architecture}
The architecture of our proposal is illustrated in Figure \ref{fig:architecture}. We apply a backbone video network  for each segment $s_i \in S$ where $S$ is composed of video segments $S = \{s_0, ..., s_{T/100}\}$, where $T$ is the number of frames in $S$. This results in video features $F = \{f_0, ..., f_{T/100}\}$. Our Therblig-Model predicts a sequence of Therbligs $\hat{t}_i$ for each $f_i \in F$. Our Action-Model takes the representations produced by the Therblig-Model along with $S$, and produces action class likelihood(s) predictions.


As the Therblig annotations $t_i$ and action annotations $a_i$ do not exhibit one-to-one overlap between their respective video sequences, the Therblig-Model and Action-Model are trained separately.


\subsubsection{Therblig-Model}
\label{section:Therblig-Model}


The primary base architecture for Therblig-Model can be any popular video architecture. Features $f_i$ are extracted from the backbone immediately prior to action classification and sent to a 2-layer GRU. Due to the lack of precise temporal alignment between the input video segments $s_i \in S$ and the Therblig annotations $t_i$, we adopt an encoder-decoder schema as follows: The hidden state of the GRU is set to $f_i$, and the network is rolled out to iteratively predict a sequence of Therbligs $\hat{t}_i = \{\hat{t}^0_i, ..., \hat{t}^{N-1}_i\}$, feeding $\vec{0}$ as the initial input, and outputs of previous hidden layers as the inputs to the decoder for subsequent timesteps. We adopt the practice of teacher forcing, where the outputs of previous hidden layers are occasionally replaced with the ground truth, with probability $p = 0.5$. After training the Therblig-Model for $30$ epochs and selecting the model instance with the top validation accuracy, we freeze the model for the training of the Action-Model. 



\subsubsection{Action-Model}
\label{section:actionlstm}

As with the Therblig-Model, the base network of the Action-Model can be any popular video architecture. Sliding windows of size $W$ is taken from input video $S$ of $T$ frames with a stride of $s$, where $s < W$ for the task of action segmentation, and $s = W$ for the task of action recognition \& anticipation. Each window is fed to the video model, producing features $f_j \in \{f_0, ..., f_{T/s}\}$ extracted immediately prior to action classification. Our action segmentation models process the sliding windows sequentially, whereas our action recognition \& anticipation base networks capture the entire video without windowing. We pair indices $i$ from $h^t_i$ and $j$ from $f_j$ by cross-referencing the closest times in video $S$ associated with $i$ and $j$. 

We adopt a temporal attention mechanism $\phi$, taking features $f_j$ as input, and producing learned attention weights $\alpha^a_i = \{\alpha^{a^0}_i, ..., \alpha^{a^{N-1}}_i\}$ over hidden layer outputs $\hat{h}_i = \{\hat{h}^{t^0}_i, ..., \hat{h}^{t^{N-1}}_i\}$. After blending the hidden states with $\alpha^a_i$, the blended features are concatenated with features $f_j$, and fed to fully connected layers to predict action class $\hat{a}$ for action recognition \& anticipation, and action classes $\hat{a}_j \in  \{a_0, ..., a_{T / s}\}$ for action segmentation.

\subsection{Therblig Rules}
\label{section:training}

\begin{figure*}[t!]
  \centering
  \includegraphics[width=.9\textwidth]{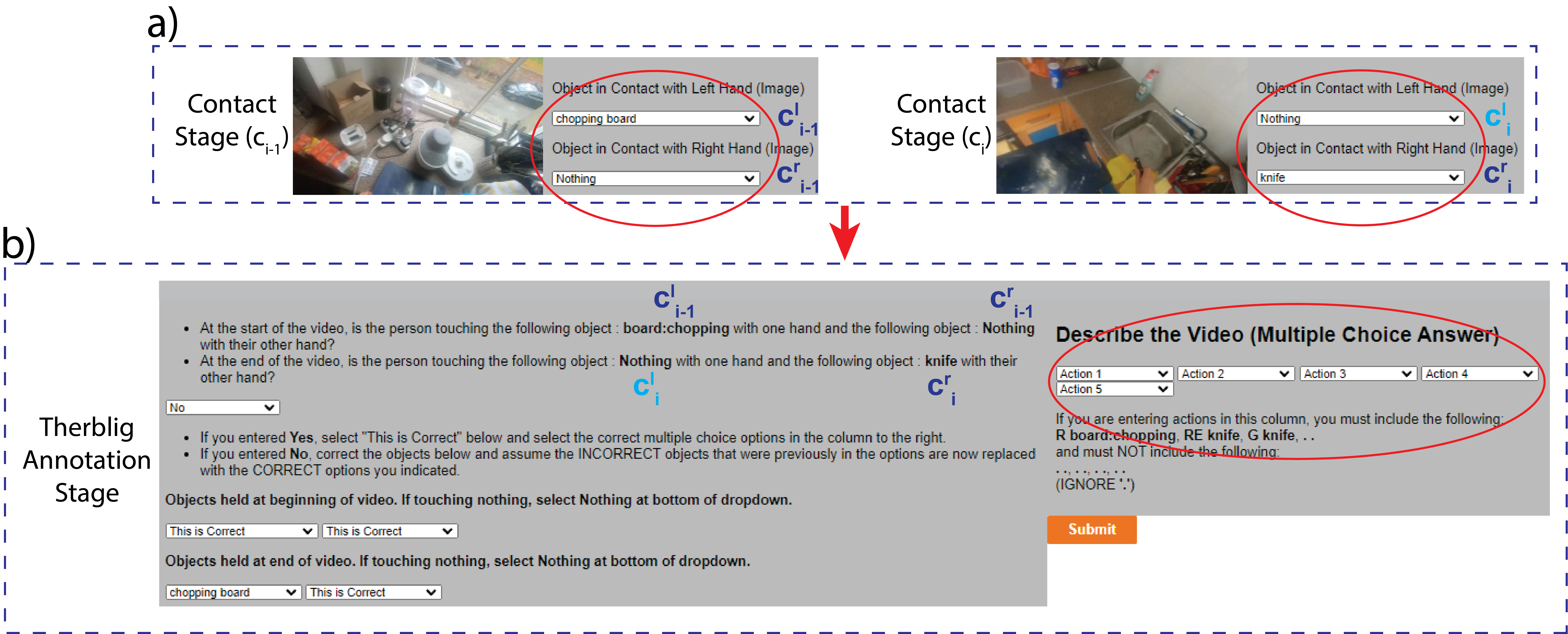}
\caption{An overview of our two-stage crowdsourcing pipeline. The \textbf{Contact Annotation Stage} provides an image (shown) paired with a video (not shown) and asks the user to indicate the objects held by the actor in the image/video via multiple choice ({\color{red} circled}). In the \textbf{Therblig Annotation Stage}, an annotator first validates the correctness of $c_i$ and $c_{i-1}$ and then produces Therblig annotations $t_i$. As can be seen above, in the Contact Annotation Stage a worker mistakenly indicated that nothing was held by the left hand for $c_i$. In the Therblig Annotation Stage, after correcting these erroneous contact annotations, a new worker annotates the sequence of Therbligs $t_i$ ({\color{red} circled}) - the multiple choice options shown to this worker have been filtered so as to achieve consistency with rules discussed in Section \ref{section:explicit}.}
 \label{fig:crowdsourcing}

\end{figure*}

Therbligs enable the introduction of contact-centered rules that 1) provide significant structure to the annotation process and 2) provide structure during training in the form of separate differentiable loss components. See Sections \ref{section:explicit} and \ref{section:diff} for more.

\subsubsection{Explicit Rules}
\label{section:explicit}


Below we enumerate the explicit commonsense rules we introduce over the Therblig ontology.


\textbf{Rule 1} The additions and subtractions of objects in contact (through grasping and releasing, respectively) produced by the Therblig sequence linking $c_i$ and $c_{i + N - 1}$ must produce object contact set $c_{i + N - 1}$ from $c_i$.

\textbf{Rule 2} Objects in contact state $c_i$ cannot be \textit{grasped} or \textit{reached} without first being \textit{released}.

\textbf{Rule 3} Objects not in contact state $c_i$ cannot be \textit{moved}, \textit{oriented}, \textit{used}, or \textit{released} without first being \textit{grasped}.

These rules structure the second stage of crowdsourcing annotations in the form of a filter over all possible annotations. Annotations violating any of Rules 1, 2 or 3 are rejected are re-sent to annotators for correction - see Section \ref{section:crowdsourcing} for details.


\subsubsection{Differentiable Rules}
\label{section:diff}
We wish to incorporate the rules discussed in Section \ref{section:explicit} into the training of the Therblig-Model. However, the rules are non-differentiable, and as such we approximate each rule with differentiable closed-form expressions.
However, the logic requires discrete representations of Therblig predictions. As the softmax activation outputs are continuous, they must be converted into their discrete equivalents. The \textit{argmax} operation is non-differentiable, and so we adopt the use of Gumbel Softmax as a differentiable alternative, applying it over the pre-softmax features of the Therblig-Model to arrive at discrete one-hot-encodings of Therbligs (${\hat{g_i}}$) while maintaining differentiability.


The rules are represented as follows:




\textbf{Rule 1} Cross Contact-Therblig Consistency Loss
\begin{equation}
  \begin{gathered}
L_C(i) = \sum_{k=0}^{N} \|c_i +  \hat{g}^k_i \beta - c_{i + 1}\| 
\end{gathered}
\end{equation}

\textbf{Rule 2} Contact Enforcement Loss 
\begin{equation}
\begin{gathered}
L_{EC}(i) = \sum_{k=0}^{N} \|a_{i,k} - \hat{g}^k_i \gamma \| \\ 
\mbox{ where } a_{i,k} = a_{i,k-1} +  \hat{g}^{k-1}_i \beta \mbox{ and } a_{i,0} = c_i
\end{gathered}
\end{equation}

\textbf{Rule 3} Non-Contact Enforcement Loss
\begin{equation}
\begin{gathered}
L_{NC}(i) = \sum_{k=0}^{N} \|a_{i,k} - \hat{g}^k_i \delta\| \\
\mbox{ where }  a_{i,k} = a_{i,k-1} + \hat{g}^{k-1}_i \beta \mbox{ and } a_{i,0} = c_i
\end{gathered}
\end{equation}

Each of $\beta$, $\gamma$, and $\delta$, correspond to a vector of size $7 \times 1$, each index of which corresponds to a single Therblig verb. The predicted Therbligs $\hat{g}_i^k$ are of dimension $|C| \times 7$, where $C$ is the set of object categories in the dataset. Contact state $c_i$ is of dimension $|C| \times 1$, taking values of $1$ for object indices corresponding to those in contact with the hand, and $0$ otherwise, based on ground truth. The $k$ variable is used to iterate over each of the $N$ elements in the Therblig sequence. The $i$ variable takes a value in $[0, T/s]$ corresponding to segments in the input clip. 


For $\beta$, values take on $1$ for \textit{grasp}, $-1$ for \textit{release}, and $0$ otherwise. These values of $\beta$ were chosen such that $ \hat{g}^k_i \beta$ reflects the addition and subtraction of objects to and from the hand. Our loss formulation in \textbf{Rule 1} penalizes predictions $\hat{g}^k_i$ that do not produce contact state $c_{i+1}$ from $c_i$.

For $\gamma$ in \textbf{Rule 2}, values take on $-1$ for \textit{reach} and \textit{grasp}, $0$ otherwise. These values of $\gamma$ were chosen such that $\hat{g}^k_i \gamma $ results in loss that penalizes \textit{reach} and \textit{grasp} Therbligs when the Therblig-derived contact state $a_{i,k}$ already contains their respective objects. 

For $\delta$ in \textbf{Rule 3}, values take on $1$ for \textit{move}, \textit{orient}, \textit{use}, or \textit{release}, and $0$ otherwise. These values of $\delta$ were chosen such that $\hat{g}^k_i \delta $ results in loss that penalizes those Therbligs when the Therblig-derived contact state $a_{i,k}$ does not contain their respective objects. 

Loss component $L_C$ of \textbf{Rule 1} measures the offset of $c_i$ and $c_{i+1}$ against $g^k_i \beta$. Loss component $L_{EC}$ of \textbf{Rule 2} iteratively compares the Therblig-derived contact state $a_{i,k}$ against $\hat{g}^k_i \gamma $. Loss component $L_{NC}$ of \textbf{Rule 3} iteratively compares $a_{i,k}$ against $\hat{g}^k_i \delta $.

We adopt each of these loss terms in addition to Categorical Cross Entropy loss $L_{CE}$ to arrive at combined loss $L = L_{CE} + L_{NC} + L_{EC} + L_{C}$. These loss terms provide meaningful constraints in guiding the learning process, and we demonstrate this finding empirically in Section \ref{section:therbligpred}.

\begin{figure}[b!]
\begin{center}
  \includegraphics[width=.35\textwidth]{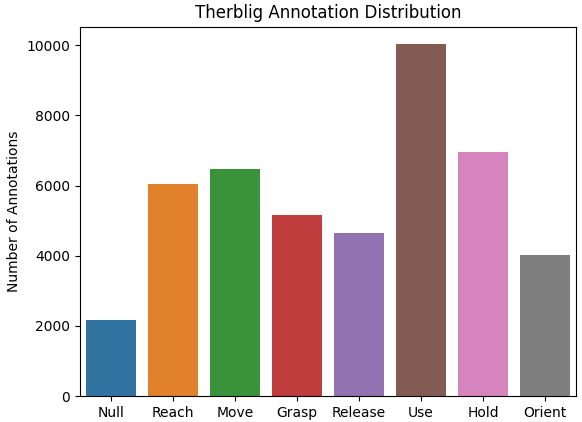}
 \end{center}
  \caption{Breakdown by frequency of Therblig verbs over all collected annotations. Null corresponds to Therblig sequences where workers indicated no hand-object activity took place.}
  \label{figure:histogram}
\end{figure}

\subsection{Crowdsourcing}
\label{section:crowdsourcing}

We crowdsource our annotations in two stages; see Figure \ref{fig:crowdsourcing}. The first stage involves the crowdsourcing of objects in contact with the hands. The annotator is shown an image, along with a slightly slowed-down video roughly $5$ seconds long. The addition of the video adds temporal context for images too difficult to annotate alone.  The annotator is asked to indicate via multiple choice the objects in contact with the actor's hands at the end of the interaction. We pool $5$ independently collected responses per image, taking the mode of the answers for consistency. The average time per assignment was $3.4$ minutes.

In the second stage, the possible Therbligs for each item in the sequence $t^j_i$ for $0 \leq j < N$ are determined by taking the cross product of the contact annotations $c_i$ and $c_{i+1}$ with $V = \{Re, M, G, R, O, U, H\}$ and filter the results such that \textit{consistency} is observed with respect to the three rules defined in Subsection \ref{section:explicit}. \textbf{We note there are $1,067$ possible Therblig $(verb, object)$ tuples; through our rules, we are able to significantly reduce the average number of multiple choice annotation possibilities to just $19$ $(verb, object)$ tuples!} We set $N$, the number of Therblig tuples per video clip, to $6$.


\begin{figure*}[t!]
  \centering
  \includegraphics[width=.9\textwidth]{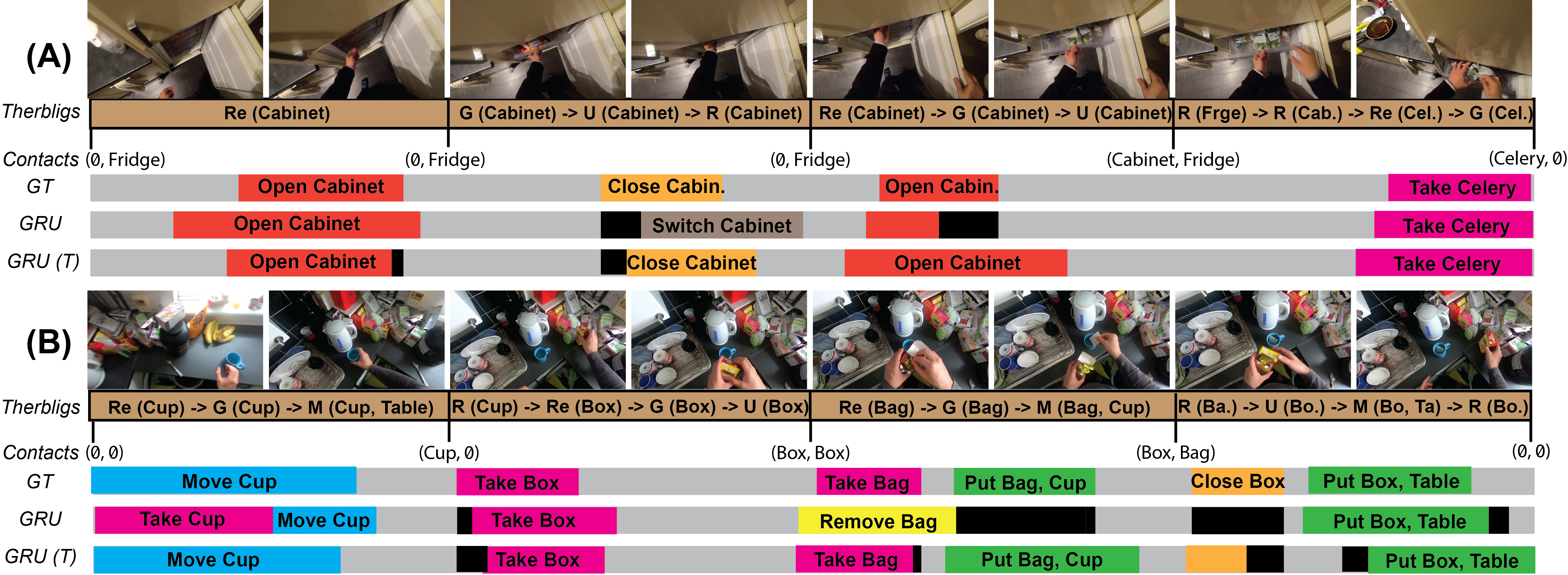}
\caption{Qualitative results for the action segmentation task showing Therblig annotations (\textit{Therbligs}), contact annotations $c_i$ (\textit{contacts}), ground truth annotations (\textit{GT}), the base GRU baseline (\textit{GRU}) and the base GRU with Therbligs (\textit{GRU (T)}). Grey colors indicate un-annotated frames. Errors in \textit{GRU} due to lack of mutual exclusivity are shown in the form of non-black contrastive colors w.r.t. \textit{GT}. Black colors indicate all other action misclassifications. Actual model predictions are at the verb-level (objects are introduced into predictions solely for purposes of illustration).}
 \label{fig:qualitative}
\end{figure*}

\section{Experiments}
\label{experiments}
Our experiments on action segmentation explore the extent to which we are able to predict Therbligs and the extent to which their incorporation benefits action segmentation\footnote{See \href{https://drive.google.com/drive/folders/1INsNVBB4r-QZUkk2lwaa3YZpti6ZIQDk?usp=sharing}{here} for an exhaustive list of videos paired with their corresponding Therblig annotations, contact state annotations, Therblig predictions and action annotations.}. We report our numbers by early stopping over validation accuracy for $3$ independent runs, reporting only the mean. All code, data and results will be released upon acceptance.

\begin{table}
\centering
\begin{tabular}{l*{6}c}
\toprule
Full Data & 50-Salads & EPIC Kitchens \\ 
\midrule
$L_{CE}$      & 22.1\%/3.19/2.25       & 9.5\%/5.68/2.491        \\
$L_{CE} + L_C$   & 23.0\%/3.15/1.96      & 9.9\%/5.209/2.124      \\
$L_{CE} + L_{EC}$ & 23.2\%/2.99/1.92  & 11.7\%/5.238/1.90 \\
$L_{CE} + L_{NC}$   & 23.1\%/2.83/1.98   & 12.7\%/\textbf{5.16}/1.91  \\
All $L$   & \textbf{25.1\%}/\textbf{2.6}/\textbf{1.70}  & \textbf{13.7\%}/5.19/\textbf{1.83} \\ 
\toprule
Low Data & & \\
\midrule
$L_{CE}$      & 12.1\%/5.91/2.53       & 7.4\%/5.37/2.36  \\
All $L$   & \textbf{16.9\%}/\textbf{5.25}/\textbf{1.96}  & \textbf{10.1\%}/\textbf{5.21}/\textbf{1.86} \\ 
\bottomrule
\end{tabular}
\vspace{1em}
\caption{Evaluation of Therblig-Model when trained over all Therblig annotations (Full Data) and when trained over a subset (Low Data). Results reported in order of : Accuracy $\uparrow$/Levenshtein Distance $\downarrow$/ Logical Consistency $\downarrow$.}
\label{table:ablations}
\end{table} 

\subsection{Datasets}
\label{section:datasets}
\textbf{EPIC Kitchens} \indent 
We choose the EPIC Kitchens 100 dataset because of the benefits the egocentric perspective provides in allowing for full view of the hands and objects in contact. We augment portions of the EPIC Kitchens dataset with densely labeled Therbligs, for a total of $14,600$ crowdsourced annotations. In accordance with reporting practices of other action segmentation approaches over EPIC Kitchens, we perform training and evaluation solely at the verb-level. See the Supplementary Materials for more.


\par
\textbf{50 Salads}  \indent We choose the 50 Salads dataset because each activity performed is strongly structured by the making and breaking of contact. We augment the entirety of the video in this dataset, for a total of $6,500$ crowdsourced annotations. See the Supplementary Materials for more.


\subsection{Therblig Prediction}
\label{section:therbligpred}
In this set of experiments, we answer the extent to which our Therblig-Model is capable of mapping video chunks $s_i$ to the sequence of Therbligs $\hat{t_i} = \{\hat{t}^0_i, ..., \hat{t}^{N-1}_i\}$.

\textbf{Metrics} \indent 
We evaluate our results, comparing $\hat{t}^j_i$ and $t^j_i$, for $0 \leq j \leq N$, $\forall i$) over the following metrics of evaluation: Element-wise accuracy and Levenshtein distance (\textbf{L}). Element-wise accuracy metrics suffer from the strict ordering requirement; and so are not reflective of sequence-level similarity. Therefore we also evaluate over Levenshtein distance, the number of edits (insertions, deletions, and swaps) to transform $\hat{t}^j_i$ into $t^j_i$. In addition, we evaluate the logical consistency of our predictions as measured by the normalized number of violations per sequence of the rules described in Section \ref{section:explicit}.

\textbf{Comparisons} \indent Table \ref{table:ablations} illustrates the results of various forms of Therblig-Model using a backbone I3D network over the EPIC Kitchens and 50 Salads datasets. \textbf{$L_{CE}$} refers to a simple, 2-layer bidirectional GRU trained solely over Categorical Cross Entropy. We train our GRU with and without each of the loss components discussed in Section \ref{section:diff}. In addition, we include results when training over \textbf{$L_{CE} + L_C + L_{EC} + L_{NC}$} and \textbf{$L_{CE}$}, but in the low-data setting where roughly $10\%$ of our annotations are trained over for both datasets, highlighting the value of the structure defined in Section \ref{section:diff}. See Table \ref{table:ablations} for results. All training and evaluation details are described in the Supplementary Materials section.

\subsection{Action Segmentation}
In this experiment we evaluate the extent to which the incorporation of Therblig-Model benefits performance in action segmentation. In accordance with \cite{huang2020improving} we perform predictions at the verb level, rather than the verb-object level.

\textbf{Metrics} \indent As action segmentation is a classic task in computer vision, we rely on the works of \cite{huang2020improving,ishikawa2021alleviating} in adopting the following evaluation metrics: frame-wise accuracy, segmental edit-score, and segmental F1-score. Frame-wise accuracy is used in all action segmentation works, whereas segmental edit-score and F1-score are most commonly used to penalize over-segmentation in particular.

\begin{table}
\centering
\begin{tabular}{l*{6}c}

\cmidrule{2-4}
& & 50 Salads & EPIC Kitchens  & \\ 
\cmidrule{2-4}
& GRU      & 62.8\%/8.45/60.9        & 39.1\%/19.3/23.2           \\
& GRU w. T & \textbf{66.9\%}/\textbf{8.21}/\textbf{65.1}   & \textbf{44.7\%}/\textbf{13.5}/\textbf{25.6}   \\
\cmidrule{2-4}

& Base MST      & 78.19\%/7.1/74.16       &  53.44\%/10.4/54.38 \\
& MST w. T & \textbf{85.25\%}/\textbf{6.7}/\textbf{81.01}  &  \textbf{58.75\%}/\textbf{9.9}/\textbf{60.97}   \\
\cmidrule{2-4}

& Base ASF      & 82.97\%/6.9/77.23       & 58.14\%/10.3/56.41           \\
& ASF w. T & \textbf{88.2\%}/\textbf{6.3}/\textbf{84.33}  & \textbf{64.03\%}/\textbf{9.5}/\textbf{61.46} \\
\cmidrule{2-4}
\end{tabular}
\caption{Action Segmentation results over EPIC Kitchens and 50 Salads datasets for frame-wise accuracy $\uparrow$/ segmental edit distance $\downarrow$/ and segmental F1-score@25 $\uparrow$. Base models (without \textbf{w. T}) are followed by base models incorporating Therbligs (\textbf{w. T}).}
\label{actseg}
\end{table}


\textbf{Comparisons} \indent Table \ref{actseg} illustrates the results of ablations of our proposed architecture over the EPIC Kitchens and 50 Salads datasets for the following base architectures of Action-Model: \textbf{GRU}, \textbf{MST} (MSTCN++), and \textbf{ASF} (ASFormer). The models (with no suffix) correspond to the mapping of raw video to framewise action labels and the models with the (\textbf{w. T}) suffix correspond to those of our proposed framework. For simplicity, we utilize the same Therblig-Model across all models with an I3D backbone, as ASFormer and MSTCN++ both use I3D features as their respective backbones. All training and evaluation details are described in the Supplementary Materials section.

\begin{table}
\centering
\begin{tabular}{l*{6}c}

\cmidrule{2-4}
& & 50 Salads & EPIC Kitchens  & \\ 
\cmidrule{2-4}
& I3D      & 64.2\%       & 65.1\%/49.7\%/39.2\%           \\
& I3D w. A     & 64.0\%       & 65.9\%/49.6\%/39.5\%           \\
& I3D w. T & \textbf{69.1\%}   & \textbf{68.2\%}/\textbf{51.9\%}/\textbf{41.6\%}  \\
\cmidrule{2-4}

& TimeSFrmr      & 73.1\%       &  62.1\%/55.4\%/41.1\% \\
& TimeSFrmr w. A     & 74.2\%       &  61.9\%/56.9\%/41.9\% \\
& TimeSFrmr w. T & \textbf{76.4\%}  & \textbf{67.6\%}/\textbf{57.0\%}/\textbf{44.0\%}   \\
\cmidrule{2-4}

& ViViT      & 72.1\%       & 62.4\%/56.0\%/43.3\%           \\
& ViViT w. A     & 72.9\%       & 64.0\%/57.2\%/44.0\%           \\
& ViViT w. T & \textbf{74.3\%}  & \textbf{66.4\%}/\textbf{56.9\%}/\textbf{45.9\%} \\
\cmidrule{2-4}

& MoViNet      & 73.9\%       & 67.9\%/52.9\%/41.1\%           \\
& MoViNet w. A     & 72.8\%       & 67.6\%/52.1\%/40.2\%           \\
& MoViNet w. T & \textbf{76.5\%}  & \textbf{69.1\%}/\textbf{55.0\%}/\textbf{43.8\%} \\
\cmidrule{2-4}

\end{tabular}
\caption{Action recognition accuracies over EPIC Kitchens and 50 Salads datasets. Results under EPIC Kitchens are provided as: verb/object/action prediction accuracies, respectively. Base models (without \textbf{w. T}) are followed by base models incorporating  action labels (\textbf{w. A}) and Therbligs (\textbf{w. T}).}
\label{rec}
\end{table}

\subsection{Action Recognition \& Anticipation}
In these experiments we evaluate the extent to which the incorporation of Therblig-Model benefits performance in action recognition \& anticipation. The Therblig-model and Action-model backbones are identical throughout all these experiments.

\textbf{Metrics} \indent We adopt accuracy as the primary metric of comparison for all action recognition \& anticipation results. We structure results over EPIC Kitchens by verb recognition accuracies, object recognition accuracies, and action recognition accuracies.

\textbf{Comparisons} \indent Tables \ref{rec} and \ref{ant} illustrate the results of ablations of our proposed architecture over the EPIC Kitchens and 50 Salads datasets for the following base architectures of Action Model: \textbf{I3D}, \textbf{TimeSFormer} (TimeSFormer-B), \textbf{ViViT} (ViViT-B/16x2) and \textbf{MoViNet} (MoViNet-A3). The models (with no suffix) correspond to the mapping of raw video to framewise action labels and the models with the (\textbf{w. T}) suffix correspond to those of our proposed framework. All models are pre-trained over Kinetics 400. For purposes of reproducibility, we describe all details in the Supplementary Materials.

\section{Discussion}
\label{Discussion}

To give better intuition we show predicted Therblig sequences alongside video of manipulation activity \href{https://drive.google.com/drive/folders/1FeE4euTMI4exkdt2MEo0TuAiQIF97KR0?usp=share_link}{here}.

We point the reader's attention towards the "Low Data" results reported in Table \ref{table:ablations}, where we observe a large increase in accuracy through the incorporation of all the rule-based loss components. We believe this validates our hypothesis that the constraints imposed by the rules play a particularly outsized role in the low-data setting, where the model would otherwise have to infer the same commonsense structure, motivating future possible directions in the few-shot domain.


\begin{table}
\centering
\begin{tabular}{l*{6}c}

\cmidrule{2-4}
& & 50 Salads & EPIC Kitchens  & \\ 
\cmidrule{2-4}
& I3D      & 39.5\%  & 31.1\%/18.3\%/10.1\%         \\
& I3D w. A     & 41.9\%  & 32.0\%/18.1\%/10.4\%         \\
& I3D w. T & \textbf{43.1\%}   & \textbf{33.0\%}/\textbf{18.9\%}/\textbf{10.9\%}  \\
\cmidrule{2-4}

& TimeSFrmr      & 48.6\%       &  31.6\%/28.2\%/13.6\% \\
& TimeSFrmr w. A     & 49.0\%       &  29.9\%/28.5\%/13.1\% \\
& TimeSFrmr w. T & \textbf{51.4\%} & \textbf{33.9\%}/\textbf{28.5\%}/\textbf{14.8\%}   \\
\cmidrule{2-4}

& ViViT      & 45.6\%       & 31.9\%/29.8\%/13.9\%           \\
& ViViT w. A & 44.9\%  & 33.0\%/\textbf{30.1\%}/14.5\% \\
& ViViT w. T & \textbf{49.1\%}  & \textbf{33.7\%}/29.4\%/\textbf{14.7\%} \\
\cmidrule{2-4}

& MoViNet      & 46.4\%       & 34.2\%/25.6\%/13.1\%           \\
& MoViNet w. A     & 47.2\%       & 34.9\%/26.0\%/13.3\%           \\
& MoViNet w. T & \textbf{48.2\%} & \textbf{36.1\%}/\textbf{26.9\%}/\textbf{14.0\%} \\
\cmidrule{2-4}

\end{tabular}
\caption{Action anticipation accuracies over EPIC Kitchens and 50 Salads datasets. Results under EPIC Kitchens are provided as: verb/object/action prediction accuracies, respectively. Base models (without \textbf{w. T}) are followed by base models incorporating action labels (\textbf{w. A}) and Therbligs (\textbf{w. T})}
\label{ant}
\end{table}

As demonstrated in Tables \ref{actseg}, \ref{rec}, and \ref{ant}, the incorporation of Therbligs results in superior performance over both the 50 Salads and EPIC Kitchens 100 datasets for all base architectures and tasks. While the accuracies of the models trained in Figure \ref{actseg} come close to matching the reported numbers in the original papers, the accuracy of the models trained with Therbligs reported in Table \ref{actseg} outperform baseline models trained by us as well as those reported in original papers. 

We point the reader's attention to the particularly pronounced role Therbligs play in improving verb recognition accuracy in the action segmentation (Table \ref{actseg}) and action recognition (Table \ref{rec}) results, even more pronounced among the Transformer models, likely due to difficulties of Transformers in capturing the fine-grained aspects of motion \cite{huang2021towards}. We also point the reader to Figure \ref{fig:qualitative} for qualitative results. When more than one action applies to any given video segment, the model incorporating Therbligs is more likely to produce interpretations of action that happen to align more closely with actions belonging in the ground truth. We hypothesize the reason to be that the incorporation of Therbligs alleviates the burden on the network of learning movement representations, and allows it to focus on contextual cues important to resolving issues of mutual exclusivity. Furthermore, the model incorporating Therbligs produces predictions with better action boundaries.

We point the reader to the Supplementary Material for more discussion.




\section{Conclusion}
\label{conclusion}


In this paper we have presented a method for mitigation of common limitations in action understanding approaches using a novel framework structured around Therbligs - a consistent, expressive, contact-centered representation of action. We demonstrate superior performance through the introduction of Therbligs over action recognition/action anticipation/action segmentation - observing an average 10.5\%/7.53\%/6.5\% improvement, respectively, over EPIC Kitchens and an average 8.9\%/6.63\%/4.8\% relative improvement, respectively, over 50 Salads. We hope the release of Therblig annotations inspires future work towards the hierarchical modeling of action, and will release all code and data upon acceptance.


%
%


\end{document}